# Global Wheat Head Detection (GWHD) dataset: a large and diverse dataset of high resolution RGB labelled images to develop and benchmark wheat head detection methods.

E. David[1,2] †, S. Madec[1,2] †, P. Sadeghi-Tehran[3], H. Aasen[4], B. Zheng[5], S. Liu[2,6], N. Kirchgessner[4], G. Ishikawa[7], K. Nagasawa[8], M.A. Badhon[9], C. Pozniak[10], B. de Solan[1], A. Hund[4], S.C. Chapman[5,11], F. Baret[2,6], I. Stavness[9]*, W. Guo[12]*

[1] Arvalis, Institut du végétal, 3, rue Joseph et Marie Hackin, 75116 Paris, France
[2] UMR1114 EMMAH, INRAE, Centre PACA, Bâtiment Climat, Domaine Saint-Paul, 228 route de l'Aérodrome, CS 40509, 84914 Avignon CEDEX, France
[3] Plant Sciences Department, Rothamsted Research, Harpenden, United Kingdom
[4] Institute of Agricultural Sciences, ETH Zurich, Universitätstrasse 2, 8092 Zurich, Switzerland
[5] CSIRO Agriculture and Food, Queensland Biosciences Precinct 306 Carmody Road, St Lucia, 4067, QLD, Australia
[6] Plant Phenomics Research Center, Nanjing Agricultural University, Nanjing, China
[7] Institute of Crop Science, National Agriculture and Food Research Organization, Japan
[8] Hokkaido Agricultural Research Center, National Agriculture and Food Research Organization, Japan
[9] Department of Computer Science, University of Saskatchewan, Canada
[10] Department of Plant Sciences, University of Saskatchewan, Canada
[11] School of Food and Agricultural Sciences, The University of Queensland, Gatton 4343, QLD, Australia
[12] Graduate School of Agricultural and Life Sciences, The University of Tokyo, 1-1-1 midori-cho, nishi-tokyo city, Tokyo, Japan

*Corresponding author. Email: ian.stavness@usask.ca; guowei@g.ecc.u-tokyo.ac.jp
†These authors contributed equally to this work

**The detection of wheat heads in plant images is an important task for estimating pertinent wheat traits including head population density and head characteristics such as health, size, maturity stage and the presence of awns. Several studies have developed methods for wheat head detection from high-resolution RGB imagery based on machine learning algorithms. However, these methods have generally been calibrated and validated on limited datasets. High variability in observational conditions, genotypic differences, development stages, and head orientation makes wheat head detection a challenge for computer vision. Further, possible blurring due to motion or wind and overlap between heads for dense populations make this task even more complex. Through a joint international collaborative effort, we have built a large, diverse and well-labelled dataset of wheat images, called the Global Wheat Head Detection (GWHD) dataset. It contains 4,700 high-resolution RGB images and 190,000 labelled wheat heads collected from several countries around the world at different growth stages with a wide range of genotypes. Guidelines for image acquisition, associating minimum metadata to respect FAIR principles and consistent head labelling methods are proposed when developing new head detection datasets. The GWHD dataset is publicly available at http://www.global-wheat.com/ and aimed at developing and benchmarking methods for wheat head detection.**

# 1. Introduction

Wheat is the most cultivated cereal crop in the world, along with rice and maize. Wheat breeding progress in the 1950s was vital for food security of emerging countries when Norman Borlaug developed semi-dwarf kinds of wheat and a complementary agronomy system (the Doubly Green Revolution), saving 300 million people from starvation [1]. However, after increasing rapidly for decades, the rate of increase in wheat yields has slowed down since the early 1990s [2], [3]. Traditional breeding still relies to a large degree on manual observation. Innovations that increase genetic gain may come from genomic selection, new high-throughput phenotyping techniques or a combination of both [4]–[7]. These techniques are essential to select important wheat traits linked to yield potential, disease resistance or adaptation to abiotic stress. Even though high throughput phenotypic data acquisition is already a reality, developing efficient and robust models to extract traits from raw data remains a significant challenge. Among all traits, wheat head density (the number of wheat heads per unit ground area) is a major yield component and is still manually evaluated in breeding trials, which is labour intensive and leads to measurement errors of around 10% [8], [9]. Thus, developing image-based methods to increase the throughput and accuracy of counting wheat heads in the field is needed to help breeders manipulate the balance between yield components (plant number, head density, grains per head, grain weight) in their breeding selections.

Thanks to increases in GPU performance and the emergence of large scale datasets [10], [11], deep learning has become the state of the art approach for many computer vision tasks, including object detection [12], instance segmentation [13], semantic segmentation [14] and image regression [15], [16]. Recently, several authors have proposed deep learning methods tailored to image-based plant phenotyping [17]–[19]. Several methods have been proposed for wheat head quantification from RGB high-resolution images. In [8],[9], the authors demonstrated the potential to detect wheat heads with a Faster-RCNN object detection network. They estimated in [8] a relative counting error of around 10% for such methods when the image resolution is controlled. In [20], the authors developed an encoder-decoder CNN model for semantic segmentation of wheat heads which outperformed traditional handcrafted computer vision techniques. [21] developed a wheat head detection and probabilistic tracking model to characterize the motion of wheat plants grown in the field.

While previous studies have tested wheat head detection methods on individual datasets, in practice these deep learning models are difficult to scale to real-life phenotyping platforms, since they are trained on limited datasets, with expected difficulties when extrapolating to new situations [8], [22], [23]. Most training datasets are limited in terms of genotypes, geographic areas and observational conditions. Wheat head morphology may significantly differ between genotypes with notable variation in head morphology, including size, inclination, colour and the presence of awns. The appearance of heads and the background canopy also change significantly from emergence to maturation due to ripening and senescence [24]. Further, planting densities and patterns vary globally across different cropping systems and environments, and wheat heads often overlap and occlude each other in fields with higher planting densities.

A common strategy for handling limited datasets is to train a CNN model on a portion of a phenotyping trial field and test it on the remaining portion of the field [25]. This is a fundamental flaw of empirical approaches against causal models: there is no theoretical guarantee that a CNN model is robust on new acquisitions. In addition, a comparison between methods from different authors requires large datasets. Unfortunately, such large and diverse phenotyping head counting datasets do not exist today because they are mainly acquired independently by single institutions, limiting the number of genotypes, the environmental and the observational conditions used to train and test the models. Further, because the labelling process is burdensome and tedious, only a small fraction of the acquired images are processed. Finally, labelling protocols may be different between institutions, which will limit model performance when trained over shared labelled datasets.

To fill the need for a large and diverse wheat head dataset with consistent labelling, we developed the Global Wheat Head Detection (GWHD) dataset that can be used to benchmark methods proposed in the computer vision community. The GWHD dataset results from the harmonization of several datasets coming from nine different institutions across seven countries and three continents. This paper details the data collection, the harmonization process across image characteristics and labelling, and the organization of a wheat head detection challenge. Finally, we discuss the issues raised while generating the dataset and propose guidelines for future contributors who wish to expand the GWHD dataset with their labelled images.

## 2. Dataset composition
### 2.1. Experiments

The labelled images comprising the GWHD dataset come from datasets collected between 2016 and 2019 by nine institutions at ten different locations (Table 1) covering genotypes from Europe, North America, Australia and Asia. These individual datasets are called "sub-datasets." They were acquired over experiments following different growing practices, with row spacing varying from 12.5 cm (ETHZ_1) to 30.5 cm (USask_1). The characteristics of the experiments are presented in Table 1. They include low sowing density (UQ_1, UTokyo_1, UTokyo_2), normal sowing density (Arvalis_1, Arvalis_2, Arvalis_3, INRAE_1, part of NAU_1) and high sowing density (RRes_1, ETHZ_1, part of NAU_1). The GWHD dataset covers a range of pedoclimatic conditions including very productive context such as the loamy soil of the Picardy area in France (Arvalis_3), silt-clay soil in mountainous conditions like the Swiss Plateau (ETHZ_1) or Alpes de Haute Provence (Arvalis_1, Arvalis_2). In the case of Arvalis_1, Arvalis_2, UQ_1, NAU_1, the experiments were designed to compare irrigated and water-stressed environments.

| Sub-dataset name | Institution | Country | Lat (°) | Long (°) | Year | Nb. of dates | Targeted stages | Row spacing (cm) | Sowing density (seeds·m-2) | Nb. of genotypes |
|---|---|---|---|---|---|---|---|---|---|---|
| UTokyo_1 | NARO & UTokyo | Japan | 36.0N | 140.0E | 2018 | 3 | Post-flowering | 15 | 186 | 66 |
| UTokyo_2 | NARO & UTokyo | Japan | 42.8N | 143.0 | 2016 | 6 | Flowering* | 12.5 | 200 | 1 |
| Arvalis_1 | Arvalis | France | 43.7N | 5.8E | 2017 | 3 | Post-flowering - Ripening | 17.5 | 300 | 20 |
| Arvalis_2 | Arvalis | France | 43.7N | 5.8E | 2019 | 1 | Post-flowering | 17.5 | 300 | 20 |
| Arvalis_3 | Arvalis | France | 49.7N | 3.0E | 2019 | 3 | Post-flowering - Ripening | 17.5 | 300 | 4 |
| INRAE_1 | INRAE | France | 43.5N | 1.5E | 2019 | 1 | Post-flowering | 16 | 300 | 7 |
| USask_1 | University of Saskatchewan | Canada | 52.1N | 106.W | 2019 | 1 | n.a | 30.5 | 250 | 16 |
| RRes_1 | Rothamsted Research | UK | 51.8N | 0.36W | 2016 | 1 | n.a | n.a | 350 | 6 |
| ETHZ_1 | ETHZ | Switzerland | 47.4N | 8.6E | 2018 | 1 | n.a | 12.5 | 400 | 354 |
| NAU_1 | Nanjing Agric. University | China | 31.6N | 119.4E | 2018 | 1 | Flowering* | 20 | 300 or 450 | 5 |
| UQ_1 | UQueensland | Australia | 27.5S | 152.3E | 2016 | 1 | Flowering - Ripening | 22 | 150 | 8 |

Table 1: Characteristics of the experiments used to acquire images for GWHD dataset.

\* images were checked carefully to ensure that heads have fully developed and flowered.

### 2.2. Image acquisition

The GWHD dataset contains RGB images captured with a wide range of ground-based phenotyping platforms and cameras (Table 2). The height of the image acquisition ranges between 1.8 m and 3 m above the ground. The camera focal length varies from 10 to 50 mm with a range of sensor sizes. The differences in camera setup lead to a range of Ground Sampling Distance (GSD) ranging from 0.10 to 0.62 mm with the half field of view along the image diagonal varying from 10° to 46°. Assuming that wheat heads are 1.5 cm in diameter, the acquired GSDs are high enough to detect heads and even awns visually. Although all images were acquired at the nadir-viewing direction, some geometric distortion may be observed for a few sub-datasets due to the different lens characteristics of the cameras used. Datasets UTokyo_1 and ETHZ_1 are particularly affected by this issue. Each institution acquired images from different platforms, including handheld, cart, mini-vehicle and gantry systems. The diversity of camera sensors and acquisition configurations resulted in a wide range of image properties, which will assist in training deep learning models to better generalize across different image acquisition conditions.

| Sub-dataset name | Vector | Camera | Focal length (mm) | Field of view (°) * | Shooting mode | Image size (pixels) | Distance to Ground (m) | GSD (mm/px) |
|---|---|---|---|---|---|---|---|---|
| UTokyo_1 | cart | Canon PowerShot G9 X Mark II | 10 | 38.15 | automatic | 5472×3648 | 1.8 | 0.43 |
| UTokyo_2 | handheld | Olympus μ850 & Sony DSC-HX90V | 7/4 | 45.5 | automatic | 3264×2488 & 4608×3456 | 1.7 | 0.6 |
| Arvalis_1 | handheld | Sony Alpha ILCE-6000 | 50 & 60 | 7.1 | automatic | 6000×4000 | 2.9 | 0.10-0.16 |
| Arvalis_2 | handheld | Sony RX0 | 7.7 | 9.99 | automatic | 800×800$^†$ | 1.8 | 0.56 |
| Arvalis_3 | handheld | Sony RX0 | 7.7 | 9.99 | automatic | 800×800$^†$ | 1.8 | 0.56 |
| INRAE_1 | handheld | Sony RX0 | 7.7 | 9.99 | automatic | 800×800$^†$ | 1.8 | 0.56 |
| USask_1 | mini-vehicle | FLIR Chameleon3 USB3 | 16 | 19.8 | fixed | 2448×2048 | 2 | 0.45 |
| RRes_1 | gantry | Prosilica GT 3300 Allied Vision | 50 | 12.8 | automatic | 3296×2472 | 3-3.5$^§$ | 0.33-0.385 |
| ETHZ_1 | gantry | Canon EOS 5D Mark II | 35 | 32.2 | fixed | 5616×3744 | 3 | 0.55 |
| NAU_1 | handheld | Sony RX0 | 24 | 16.9 | automatic | 4800×3200 | 2 | 0.21 |
| UQ_1 | handheld | Canon 550D | 55 | 17.3 | automatic | 5184×3456 | 2 | 0.2 |

Table 2: Image characteristics of the sub-datasets comprising the GWHD dataset. All cameras looked vertically downward.

\* The field of view is measured diagonally. The reported measure is the half-angle.

$^†$ Original images were cropped, and a sub-image of size 800x800 was extracted from the central area

$^§$ The camera was positioned perpendicular to the ground and automatically adjusted to ensure a 2.2 m distance was maintained between the camera and canopy.

### 2.3. Data harmonization

An important aspect of assembling the GWHD dataset was harmonizing the various sub-datasets (Figure 1). A manual inspection of images was first conducted to ensure that they could be well interpreted. Images acquired at too early of a growth stage were removed when heads were not clearly visible (Figure 2d). Most of the images were also acquired before the appearance of head senescence since heads tend to overlap when the stems start to bend at this stage.

Object scale, i.e. the size of the object in pixels, is an important factor in the design of object detection methods [8]. Object scale depends on the size (mm) of the object and on the resolution of the image. Wheat head dimensions may vary across genotypes and growth conditions, but are generally around 1.5 cm in diameter and 10 cm in length. The actual image resolution, at the level of wheat heads, varied significantly between sub-datasets: the GSD varies by a factor of 5 (Table 1) while the actual resolution at the head level also depends on canopy height and the panoramic effect of the camera. The panoramic effect will be much larger when images were acquired too close to the canopy. Images were therefore rescaled to keep more similar resolution at the head level. Bilinear interpolation was used to up- or down-sample the original images. The scaling factor applied to each sub-dataset is displayed in Table 3.

Most deep learning algorithms are trained with square-sized image patches. When the original images were cropped into square patches, the size of the patches was selected to reduce the chance that heads would cross the edges of the patches and be partly cut off. Images were therefore split into 1024 x 1024 squared patches containing roughly 20 to 60 heads each, with only a few heads crossing the patch edges. The number of patches per original image varied from 1 to 6 depending on the sub-dataset (Table 3). These squared patches will be termed "images" for the remainder of the paper.

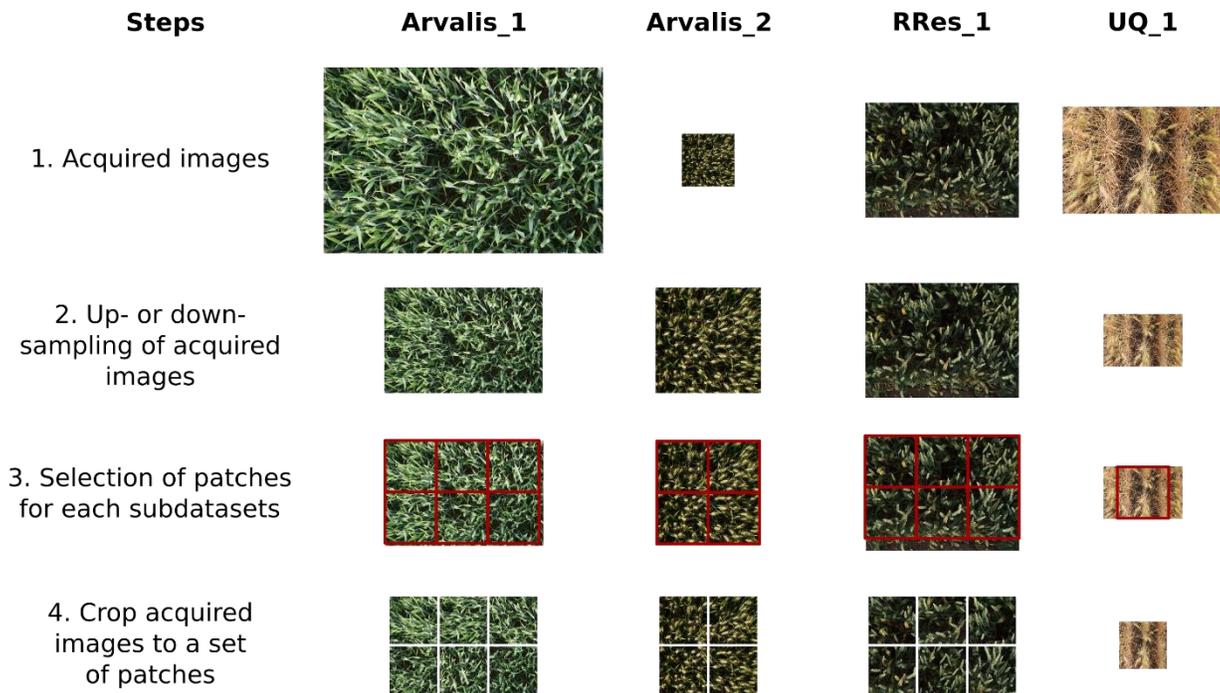
Figure 1: Overview of the harmonization process conducted. Images were first rescaled using bilinear interpolation up- or -down-sampling techniques. Then, the rescaled images were split into 1024 x 1024 squared patches.

### 2.4. Labelling

A web-based labelling platform was developed to handle the evaluation and labelling of the shared sub-datasets using the coco annotator (https://github.com/jsbroks/coco-annotator; [26]). The platform hosts all the tools required to label objects. In addition, it also grants simultaneous access to different users, thus allowing contributions from all institutions. Wheat heads were interactively labelled by drawing bounding boxes that contained all the pixels of the head. Labelling is difficult if heads are not clearly visible, i.e. if they are masked by leaves or other heads. We did not label partly hidden heads unless at least one spikelet was visible. This was mostly the case for images acquired at an early stage when heads were not fully emerged. Overlap among heads was more frequently observed when the images were acquired using a camera with a wide field of view as in UTokyo_2 or ETHZ_1. These overlaps occurred mainly towards the borders of the images with a more oblique view angle. When the bounding box was too large to include the awns, it was restricted to the head only (Figure 2a). Further, heads cropped at the image edges were labelled only if more than 30% of their basal part was visible (Figure 2e).

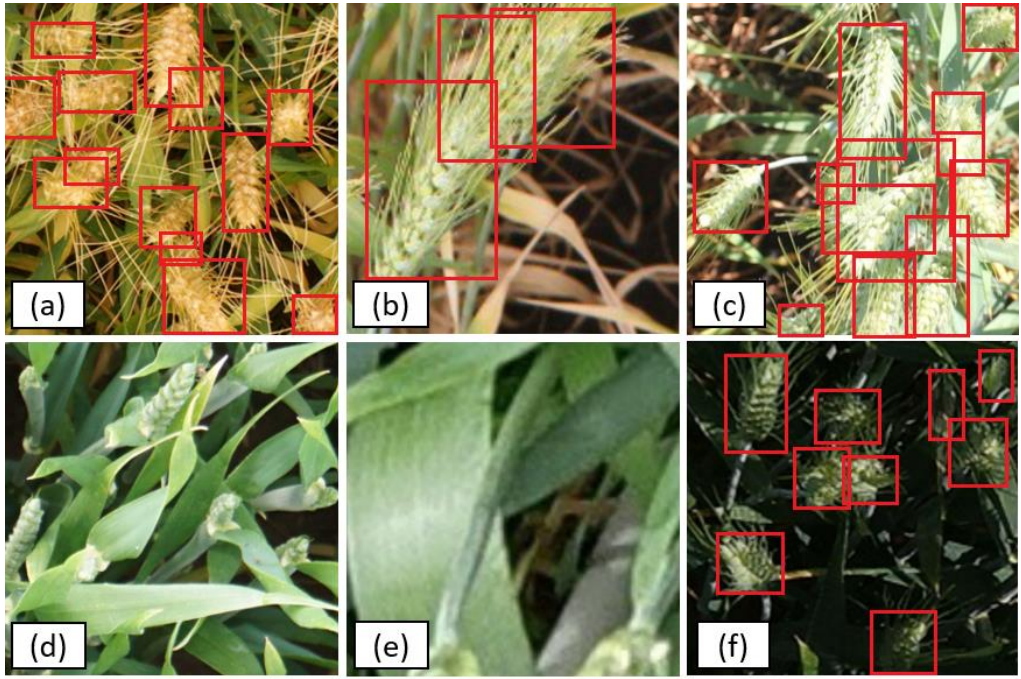

Figure 2: Examples of wheat heads difficult to label. These examples are zoomed-in views from images contained in the dataset, with different zoom factors. It includes overlapping heads (a-c), heads at emergence (d), heads that are partly cut at the border of the image (e), and images with a low illumination (f). Note that the image (d) was removed from the dataset because of the ambiguity of heads at emergence. Wheat heads in the image (e) were not labelled because less than 30% of their basal part is visible, as defined in 2.4.

Several institutions had already labelled their sub-datasets. For the datasets not labelled, we used a "weakly supervised deep learning framework" [27] to label images efficiently for these sub-datasets. A YoloV3 model [28] was trained over UTokyo_1 and Arvalis_1 sub-datasets and then applied to the un-labelled sub-datasets. Boxes with an associated confidence score greater than and equal to 0.5 were retained and proposed to the user for correction. This semi-automatic active learning increased the throughput of the labelling process by a factor of four as compared to a fully manual process. The process is detailed in Figure S1.

This first labelling result was then reviewed by two individuals independent from the sub-datasets institution. When large discrepancies between reviewers were observed, another labelling and reviewing round was initiated. Approximately 20 individuals contributed to this labelling effort. This collaborative process and repeated reviews ensure a high level of accuracy and consistency across the sub-datasets.

## 3. Description of the dataset
### 3.1. General statistics

The GWHD dataset represents 4,698 squared patches extracted from the 2219 original high-resolution RGB images acquired across the 11 sub-datasets (Table 3). It represents 188,445 labelled heads which average 40 heads per image in good agreement with the 20 to 60 targeted heads per image. However, the distribution among and within sub-datasets is relatively broad (Figure 3a). We included about 100 images that contain no heads to represent in-field capturing conditions and add difficulty for benchmarking. Few images contain more than 100 heads with a maximum of 120 heads. Multiple peaks corresponding to the several sub-datasets (Figure 3b) can be observed due mainly to variations in head density that depends on genotypes and environmental conditions. The size of the bounding boxes around the heads shows a slightly skewed gaussian distribution with a median typical dimension of 77 pixels (Figure 3b). The typical dimension is computed as the square root of the area. It corresponds well to the targeted scale, i.e. 1.5 cm x 10 cm approximate head size with an average resolution close to 0.4 mm/pixel which represents a typical dimension of 97 pixels per head, although the simple horizontal area projected does not correspond exactly to the viewing geometry of the RGB cameras. The harmonization of object scale across sub-datasets can be further confirmed visually in Figure 4.

| Sub-dataset name | Nb. of acquired images | Nb. of patch per image | Original GSD (mm) | Sampling factor | Used GSD (mm) | Nb. of labelled images | Nb. of labelled heads | Average nb. of heads/ images |
|---|---|---|---|---|---|---|---|---|
| UTokyo_1 | 994 | 1 | 0.43 | 1 | 0.43 | 994 | 29174 | 29 |
| UTokyo_2 | 30 | 4 | 0.6 | 2 | 0.3 | 120 | 3263 | 27 |
| Arvalis_1 | 239 | 6 | 0.23 | 0.5 | 0.46 | 1055* | 45716 | 43 |
| Arvalis_2 | 51 | 4 | 0.56 | 2 | 0.28 | 204 | 4179 | 20 |
| Arvalis_3 | 152 | 4 | 0.56 | 2 | 0.28 | 608 | 16665 | 27 |
| INRAE_1 | 44 | 4 | 0.56 | 2 | 0.28 | 176 | 3701 | 21 |
| USask_1 | 100 | 2 | 0.45 | 1 | 0.45 | 200 | 5737 | 29 |
| RRes_1 | 72 | 6 | 0.33 | 1 | 0.33 | 432 | 20236 | 47 |
| ETHZ_1 | 375 | 2 | 0.55 | 1 | 0.55 | 747* | 51489 | 69 |
| NAU_1 | 20 | 1 | 0.21 | 1 | 0.21 | 20 | 1250 | 63 |
| UQ_1 | 142 | 1 | 0.2 | 0.5 | 0.4 | 142 | 7035 | 50 |
| total | 2219 | - | - | - | - | 4698 | 188445 | - |

Table 3: Statistics for each component of the Global wheat head Detection

*some labelled images have been removed during the labelling process

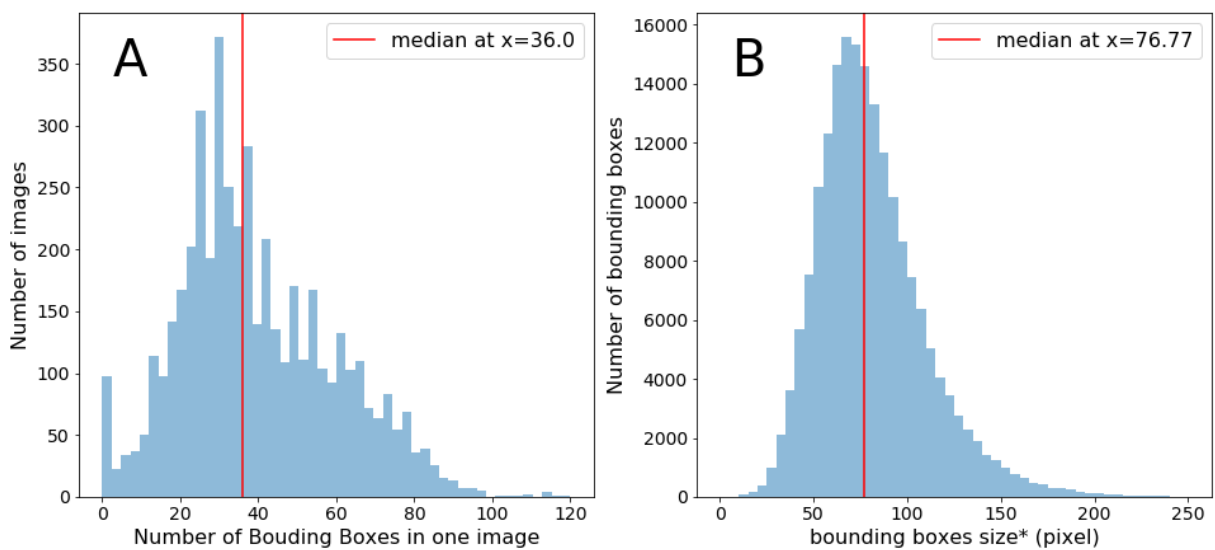

Figure 3: Distribution of the number of bounding boxes per image (a) and bounding boxes size* (b) in the GWHD Dataset.

*the bounding box size is defined as the square root of the bounding box area in pixel.

### 3.2. Diversity of sampled genotypes, environments, and developmental stages

The diversity of acquisition conditions sampled by the GWHD dataset is well illustrated in Figure 4: illumination conditions are variable, with a wide range of heads and background appearance. Further, we observe variability in head orientation and view directions, from an almost nadir direction up to a mostly oblique direction as in the case of ETHZ_1 (Figure 4). A selection of bounding boxes extracted from the several sub-datasets (Figure 5) shows a variation of bounding-box area and aspect ratio, depending on the head orientation and viewing direction. A large diversity of head appearance is observed, with variation in the presence of awns and awn size, head colour, and blurriness. In addition, a few heads were cut off when the bounding box crossed the edge of the image.

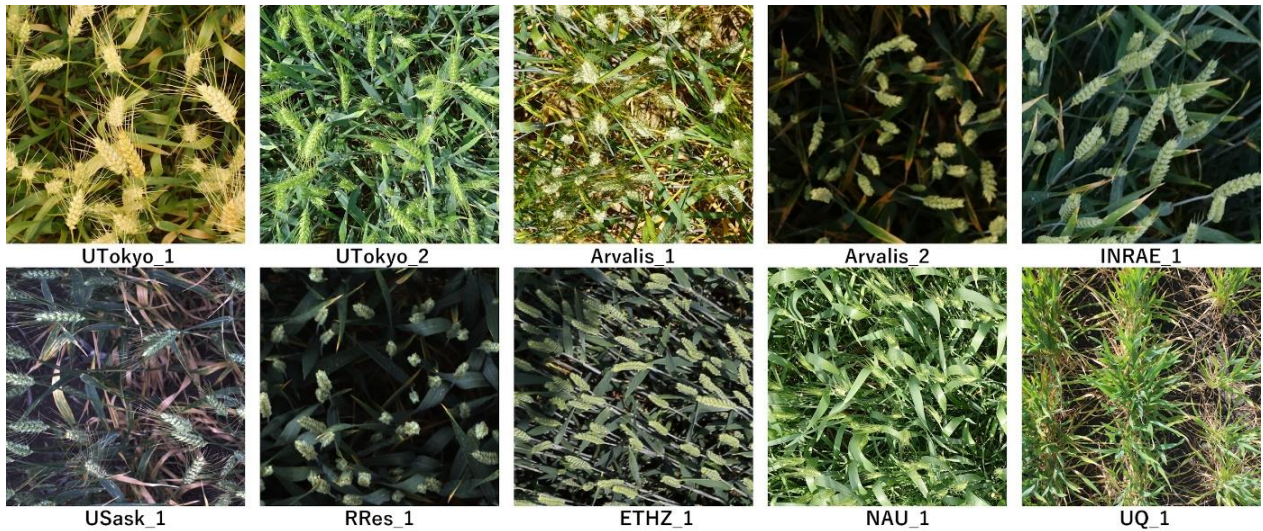

Figure 4: Example of images from different acquisition sites after cropping and rescaling.

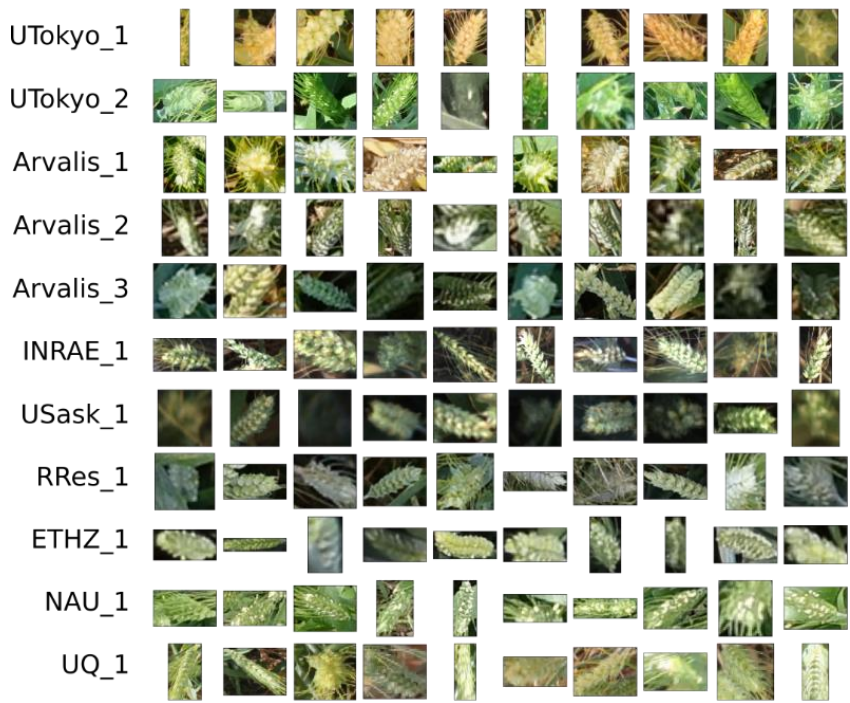

Figure 5: A selection of bounding boxes for each sub-dataset. The same size of pixels is used across all the bounding boxes displayed.

## 3.3. Comparison to other datasets

Several open-source datasets have already been proposed in the plant phenotyping community. The CVPPP datasets [29] have been widely used for rosette leaf counting and instance segmentation. The KOMATSUNA dataset also includes segmented rosette leaves, but in time-lapse videos [30]. The Nottingham ACID Wheat dataset includes wheat head images captured in a controlled environment with individual spikelets annotated [31]. However, comparatively few open-source datasets include images from outdoor field contexts, which are critical for the practical application of phenotyping in crop breeding and farming. A few datasets have been published for weed classification [32][33]. The GrassClover dataset includes images of forage fields and semantic segmentation labels for grass, clover and weed vegetation types [34]. Datasets for counting sorghum [27][35] and wheat heads [36] have also been published with dot annotations.

In terms of phenotyping datasets for object detection, our GWHD dataset is currently the largest open labelled dataset freely available for object detection for field plant phenotyping. MinneApple [37] is the only comparable dataset in terms of diversity in the field of phenotyping but proposes fewer images and less diversity in terms of location. Other datasets like MS COCO [38] or Open Images V4 [39] are much larger and sample many more object types for a wide range of other applications. The corresponding images usually contain fewer objects, typically less than ten per image (Figure 6). However, some specific datasets like PUCPR [40], CARPK [41], SKU-110K [42] are tailored to the problem of detecting objects (e.g., cars, products) in dense contexts. They have a much higher object density than the GWHD dataset, but with fewer images for PUCPR and CARPK, while SKU-110 contains more images than our GWHD dataset (Figure 6). The high occurrence of overlapping and occluded objects is unique to the GWHD dataset. This makes labelling and detection more challenging, especially compared to SKU-110K, which does not seem to present any occlusion. Finally, wheat heads are complex objects that have a wide variability of appearance as demonstrated previously, surrounded by a very diverse background which would constitute a more difficult problem than detecting cars or densely packed products on store shelves.

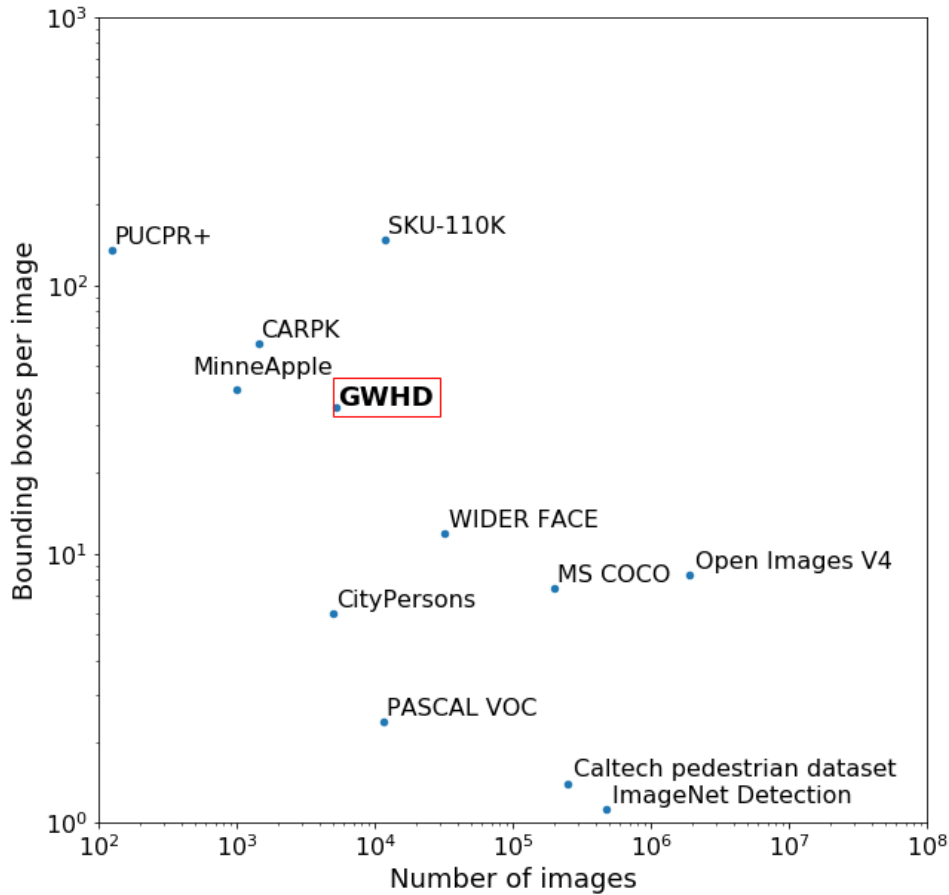

Figure 6: Comparison of GWHD dataset with other object detection datasets. Both axes are in log-scale.

## 4. Target use case: Wheat Head Detection challenge

The main goal of the dataset is to contribute to solving the challenging problem of wheat head detection from RGB high-resolution images. An open machine learning competition will be held from May to August 2020 to benchmark wheat head detection methods using the GWHD dataset for training and testing (http://www.global-wheat.com/2020-challenge/).

### 4.1. Split between training and testing datasets

In machine learning studies, it is common to randomly split a dataset into training and testing samples. However, for the GWHD competition, we specifically aim to test the performance of the method for unseen genotypes, environments, and observational conditions. Therefore, we grouped all images from Europe and North America as the training dataset, which covers enough diversity to train a generic wheat head detection model. This training dataset corresponds to 3422 images representing 73% of the whole GWHD dataset images. The test data set includes all the images from Australia, Japan, and China, representing 1276 images to evaluate model performance, including robustness against unseen images.

## 4.2. Evaluation Metrics

The choice of bounding boxes as labels in the GWHD dataset allows it to be used for object detection. The mean average precision computed from the true and false positives is usually used to quantify performance in object detection tasks. A true positive corresponds to a predicted bounding box with an intersection over union (IoU) greater than and equal to 0.5 with the closest labelled bounding box. A false positive corresponds to a predicted bounding box with an IoU strictly lower than 0.5 with the closest labelled bounding box. In the case of two predicted boxes with an IoU greater than or equal to 0.5 on the same bounding box, the most confident one is considered as a true positive and the other as a false positive. The mean Average Precision noted as mAP@0.5 is the considered metric for evaluating the localization performance. Detection of individual wheat heads is required for characterizing their size, inclination, colour, or health. However, the number of wheat heads per image is also a highly desired trait. Future competitions using the GWHD dataset could focus on wheat head counting with metrics such as the Root Mean Square Error (RMSE), relative RMSE (rRMSE), and Coefficient of Determination ($R^2$) to quantify the performance of object counting methods.

## 4.3. Baseline method

To set a baseline detection accuracy for the GWHD dataset, we provide results based on a standard object detection method. We trained a two-stage detector, Faster-RCNN [12], with a ResNet34 and ResNet50 as the backbone. Faster-RCNN is one of the most popular object detection models and used in Madec et al. [8]. ResNet34 is used along with ResNet50 because it is less prone to overfitting and faster to train. Due to memory constraints, the input size was set to 512x512 pixels. We randomly sampled ten patches of size 512 x 512 pixels for each image in the training dataset resulting in a training dataset composed of 34220 patches. We predicted on a set of overlapping patches of size 512 x 512 pixels regularly extracted from the test images of size 1024 x 1024 pixels and then merged the results. After 10 epochs, representing 342200 iterations in total, the best model is obtained at epoch 3 for both backbones. It yielded a mAP@0.5 of 0.77 and a mean RMSE of 12.63 wheat heads per image which corresponds to rRMSE=39%. The coefficient of regression is 0.57. All results are provided in Figure S2. The relatively poor performance of a standard object detection network on the GWHD dataset provides an opportunity for substantial future improvement with novel methods . The GWHD competition is expected to instigate new wheat head detection approaches that will provide more accurate results.

## 5. Discussion
### 5.1. Image acquisition recommendations

To successfully detect wheat heads, they should be fully emerged and clearly visible within the images, with minimum overlap among heads and leaves. For some genotypes and environmental conditions, we observed that the wheat stems tend to bend for the latest grain filling stages, which increases the overlap between heads. Conversely, for the stages between heading and flowering, some heads are not yet fully emerged and are therefore difficult to see. Therefore, we recommend acquiring images immediately after flowering when the wheat heads have fully emerged and are still upright in the field.

For image acquisition, a near nadir viewing direction is recommended to limit the overlap between heads, especially in the case of populations with high head density. Likewise, a narrow field of view is preferred. However, a narrow field of view may result in a small image footprint when the camera is positioned at a height close to the top of the canopy. Therefore we recommend increasing the camera height to get a larger sampled area and reduce the number of heads that will be cropped at the edge of the image. The size of the sampled area is important when head identification is used for estimating the head population density. The minimum sampled area should be that of our squared patch, i.e. 1024x1024 pixels of 0.4 mm/pixel which corresponds to an area of about 40 cm². To achieve this sampled area, while maintaining a narrow field of view of ±15°, the distance between the camera and the top of the canopy should be around 1.0 m. However, a larger sampling area is preferable for head population density estimation, where at least 100 cm² should be sampled to account for possible heterogeneity across rows. This would be achieved with a 2.5 m distance between the camera and the top of the canopy.

When estimating wheat head density, i.e. the number of heads per unit ground area, accurate knowledge of the sampled area is critical. The non-parallel geometry of image acquisition, with significative "fisheye" lens distortion effects, induces uncertainty about the sampled area. Even for our typical case with limited distortion effects (±15° field of view), for an image acquired at 2.5 m from the top of the canopy, an error of 10 cm in canopy height estimation induces 8% error in the sampled area, which directly transfers to the head density measurement. Further, the definition of the reference height at which to compute the sampled area is still an open question, because within a population of wheat plants the heights of the heads can vary by more than 25 cm, which induces a 21% difference in the sampled area between the lowest and highest head. Further work should investigate this important question.

Finally, our experience suggests that using a sub-millimetre resolution at the top of the canopy is required for efficient head detection. However, the optimal resolution is yet to be defined. Previous work [8] recommended 0.3 mm GSD, while the GWHD dataset includes GSD ranging from 0.28 to 0.55 mm. Further work should investigate this important aspect of wheat imaging, particularly regarding the possibility to use UAV observations for head density estimation in large wheat breeding experiments.

## 5.2. Minimum information associated with the sub-datasets and FAIR principles

The FAIR principles (Findable, Accessible, Interoperable and Reusable [43]) should be applied to the images that populate the GWHD dataset. A minimum set of metadata should be associated with each image as proposed in [44] to verify the FAIR principles. The lack of metadata was an issue for precise data harmonization and is limiting factor for further data interpretation [45] and possible meta-analysis. Therefore, we recommend attaching a minimum set of information to each image and sub-dataset. In our case, a sub-dataset generally corresponds to an image acquisition session, i.e. a series of images acquired over the same experiment on the same date and with the same camera. The experiment metadata are all the metadata related to agronomic characteristics of the session; the acquisition metadata are all the metadata related to the camera and acquisition vehicle used. Both can be defined at the session level and the image level. Our recommendations are summarized in Table 4. We encourage attaching more metadata such as camera settings (model, white balance correction, et al.) when possible because it adds context for further data reuse.

|  | Session level | Image level |
| --- | --- | --- |
| Experiment metadata | Name of the experiment (PUID) [†]<br>Name of institution<br>GPS coordinates (°)<br>Email address of the contact person<br>Date of the session (yyyymmdd)<br>Wheat species (durum, aestivum …)*<br>Development stage / ripening stage* | Microplot id<br>Row spacing<br>Sowing density<br>Name of the genotype (or any identifier) [†]<br>presence or not of awns. |
| Acquisition metadata | **Vector characteristics**:<br>Name<br>Type (handheld, cart, phenomobile, gantry, UAV)<br>Sampling procedure<br>Distance to the ground (m)*<br>**Camera characteristics,**<br>Model,<br>Focal length of the lens (mm),<br>Size of the pixel at the sensor matrix (μm)<br>Sensor dimensions (pixels x pixels), | Camera aperture<br>Shutter speed<br>ISO<br>Distance from camera to canopy (m) [‡]<br>Position of the image in the microplot [§] |

Table 4: The minimum metadata that should be associated with images of wheat heads.

* this may be alternatively reported at the image level if it is variable within a session
[†] persistent unique identifier (PUID). This may be a DOI as for genetic resources regulated under the on Plant Genetic Resources for Food and Agriculture (https://ssl.fao.org/glis) or any other identifier including the information of the maintainer of the genetic material, ripening stage
[‡] The distance between camera and canopy is an essential piece of information to harmonize dataset and calculate the density and should be carefully monitored during an acquisition.
§ In case of multiple images over the same microplot.

## 5.3. Need for GWHD expansion

The innovative and unique aspect of the GWHD dataset is the significant number of contributors from around the world, resulting in a large diversity across images. However, the diversity within each continent and environmental conditions are not well covered by the current dataset: more than 68% of the images within the GWHD dataset come from Europe and 43% from France. Further, some regions are currently missing, including Africa, Latin America, and the Middle East. As future work, we hope to expand the GWHD dataset in order to get a more comprehensive dataset. Therefore, we invite potential contributors to complement the GWHD dataset with their sub-datasets. The proposed guidelines for image acquisition and the associated metadata should be followed to keep a high level of consistency and respect the FAIR principles. We encourage potential contributors to contact the corresponding authors through www.global-wheat.com. We also plan to extend the GWHD dataset in the future for classification and segmentation tasks at the wheat head level, for instance, the size of the wheat head, or flowering state. This expansion would require an update of the current labels.

## 6. Conclusion

Object detection methods for localizing and identifying wheat heads in images are useful for estimating head density in wheat populations. Head detection may also be considered as a first step in the search for additional wheat traits, including the spatial distribution between rows, the presence of awns, size, inclination, colour, grain filling stage, and health. These traits may prove useful for wheat breeders and some may help farmers to better manage their crops.

In order to improve the accuracy and reliability of wheat head detection and localization, we have assembled the Global Wheat Head Detection dataset — an extensive and diverse dataset of wheat head images. It is designed to develop, and benchmark head detection methods proposed by the community. It represents a large collaborative international effort. An important contribution gained through the compilation of diverse sub-datasets was to propose guidelines for image acquisition, minimum metadata to respect the FAIR principles and guidelines and tools for labelling wheat heads. We hope that these guidelines will enable practitioners to expand the GWHD dataset in the future with additional sub-datasets that represent even more genotypic and environmental diversity. The GWHD dataset has been proposed together with an open research competition to find more accurate and robust methods for wheat head detection across the wide range of wheat growing regions around the world. The solutions proposed in the competition will be made open-source and shared with the plant phenotyping community.

## Acknowledgements

The French team received support from ANRT for the CIFRE grant of Etienne David, co-funded by Arvalis. The study was partly supported by several projects including ANR PHENOME, ANR BREEDWHEAT, CASDAR LITERAL and FSOV "Plastix". Many thanks to the people who annotated the French datasets, including Frederic Venault, Xiuliang Jin, Mario Serouard, Ilias Sarbout, Carole Gigot, Eloïse Issert, Elise Lepage.

The Japanese team received support from JST CREST Grant Number JPMJCR16O3, JPMJCR16O2, JPMJCR1512 and MAFF Smart-breeding system for Innovative Agriculture (BAC1003), Japan. Many thanks to the people who annotated the Japanese dataset, including Kozue Wada, Masanori Ishii, Ryuuichi kanzaki, Sayoko Ishibashi, Sumiko Kaneko.

The Canadian team received funding from the Plant Phenotyping and Imaging Research Center through a grant from the Canada First Research Excellence Fund. Many thanks to Steve Shirtliffe, Scott Noble, Tyrone Keep, Keith Halco, and Craig Gavelin for managing the field site and collecting images.

Rothamsted Research received support from the Biotechnology and Biological Sciences Research Council (BBSRC) of the United Kingdom as part of the Designing Future Wheat (BB/P016855/1) project. We are also thankful to Prof Malcolm J Hawkesford, who leads the DFW project and Dr Nicolas Virlet for conducting the experiment at Rothamsted Research.

The Gatton, Australia dataset was collected on a field trial conducted by CSIRO and UQ, with trial conduct and measurements partly funded by the Grains Research and Development Corporation (GRDC) in project CSP00179. A new GRDC project involves several of the authors and supports their contribution to this paper.

The dataset collected in China was supported by the Program for High-Level Talents Introduction of Nanjing Agricultural University (440—804005). Many thanks to Jie Zhou and many volunteers from Nanjing Agricultural University to accomplish the annotation.

The data set collection at ETHZ was supported by Prof. Achim Walter, who leads the Crop Science group. Many thanks to Kevin Keller for the initial preparation of the ETHZ dataset, and Lara Wyser, Ramon Winterberg, Damian Käch, Marius Hodel and Mario Serouard (INRAE) for annotation of the ETHZ dataset and to Brigita Herzog and Hansueli Zellweger for crop husbandry.

**Author contributions:**
E.D., S.M., B.S, F.B. organized field experiment and data collection for France dataset. P.S.T organized field experiment and data collection for U.K. dataset. H.A., N.K, A.H. organized field experiment and data collection for Switzerland dataset. G.I., K.N., W.G. organized field experiment and data collection for Japan dataset. S.L., F.B. organized field experiment and data collection for China dataset. C.P., M.B., I.S. organized field experiment and data collection for Canada dataset. B.Z., S.C.C organized field experiment and data collection for Australia dataset. E.D and S.M harmonized the sub-datasets. W.G, E.D. and S.M. built the initial Wheat Head Detection model and conducted pre-labelling process. E.D. administered the labelling platform, and all authors contributed to data labelling and quality check. E.D. built the baseline model for the competition. E.D. and S.M. wrote the first draft of the manuscript; they contributed equally to this work. All authors gave input and approved the final version.

**Funding:** integrate into acknowledgements.

**Competing interests:** The authors declare that there is no conflict of interest regarding the publication of this article.

**Data Availability:** The GWHD Dataset will be publicly available online (www.global-wheat.com) under the MIT license.

## Supplementary Materials

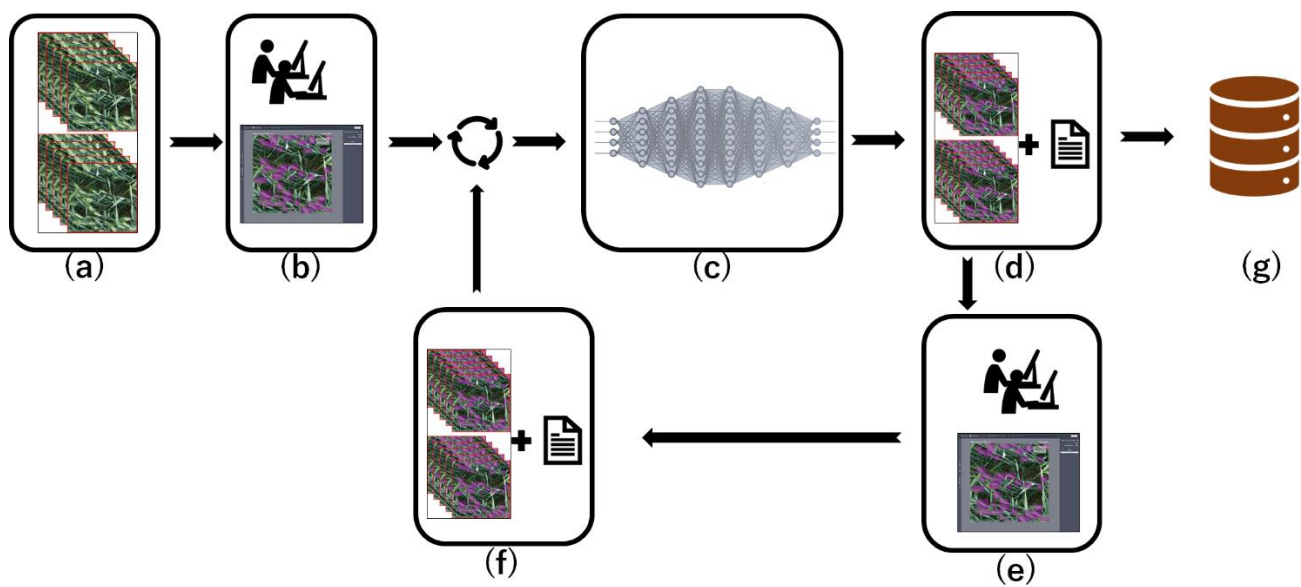

Figure S1: the proposed "weakly supervised deep learning framework" to pre-label images efficiently.

(a) Original image input.
(b) Labelling work for randomly selected original images.
(c) Train an initial model with labelled images from (b).
(d) Apply the initial model on original images left.
(e) Feed the generated labels and the image into the image annotator app for validating the bounding box locations and corrections by a human annotator.
(f) Add corrected labels and the image to data pool (b)
(g) Acquire the final annotated dataset after iterating the process (a)~(f)~(d) until reached to desired performance as determined by evaluating the trained model at the end of every iteration.

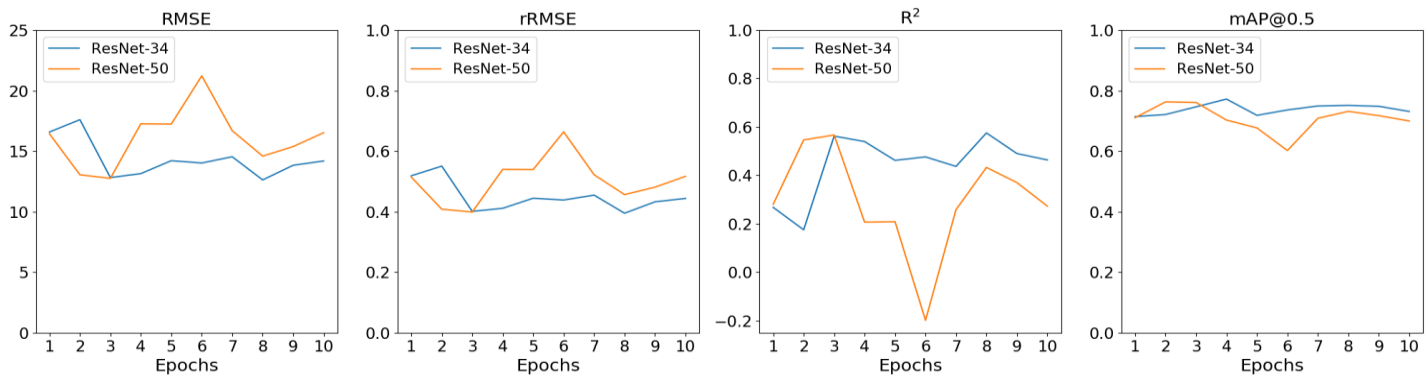

Figure S2: Epoch-wise results (RMSE, rRMSE, $R^2$, mAP@0.5) of Faster-RCNN baseline with ResNet34 and ResNet50. The best model is obtained at epoch 3 for both backbones.

.